# Optimizing LLMs for Resource-Constrained Environments: A Survey of Model Compression Techniques

Sanjay Surendranath Girija
sanjaysg@google.com

Lakshit Arora
lakshit@google.com

Aman Raj
amanraj@google.com

Shashank Kapoor
shashaankkapoor@google.com

Dipen Pradhan
dipenp@google.com

Ankit Shetgaonkar
ankiit@google.com

Google

*Abstract*— Large Language Models (LLMs) have revolutionized many areas of artificial intelligence (AI), but their substantial resource requirements limit their deployment on mobile and edge devices. This survey paper provides a comprehensive overview of techniques for compressing LLMs to enable efficient inference in resource-constrained environments. We examine three primary approaches: Knowledge Distillation, Model Quantization, and Model Pruning. For each technique, we discuss the underlying principles, present different variants, and provide examples of successful applications. We also briefly discuss complementary techniques such as mixture-of-experts and early-exit strategies. Finally, we highlight promising future directions, aiming to provide a valuable resource for both researchers and practitioners seeking to optimize LLMs for edge deployment.

## I. Introduction

LLMs have emerged as groundbreaking technologies, significantly advancing the field of AI in recent years. These models display remarkable generalization capabilities beyond their initial training objectives and tackle a wide variety of tasks [1], [2]. Beyond natural language processing and understanding, LLMs display reasoning capabilities, multimodal understanding, and generative abilities. They have been widely used in various domains like information retrieval [53], content generation [5], [6], scientific discovery [8], healthcare [7] and education [6].

While LLMs have the potential for transforming several domains, they are typically very resource-intensive to train and serve. LLMs have millions or billions of parameters which means that they require dedicated resources (machines, GPUs, TPUs, RAM) for training and inference. For example, LLaMa 3 405B does not fit in a single machine with 8 Nvidia H100 GPUs (800 GB of combined memory) and needs to be split across two machines for inference [4]. DeepSeek-V3 [62] in 16-bit precision requires 1.34 TB of GPU memory. Even the LLaMa 7B model, in 16-bit precision, requires 14 GB of GPU memory for the parameters and an additional 2 GB of memory for the Key-Value cache depending on the configuration [3]. The inference cost of these models makes them prohibitively expensive to run on mobile and edge devices.

To bridge the gap of using powerful models on resource constrained environments and edge devices, researchers have developed model compression techniques that reduce model size and inference cost while preserving accuracy. This paper examines three major techniques for efficient LLM compression: Knowledge Distillation (KD), Model Quantization, and Pruning, along with their variants and practical applications. We also briefly discuss other techniques to improve model efficiency such as mixture-of-experts and early exit strategies. While existing work often focuses on one or two compression techniques in great detail [9], [10], [13], [32], [55], this paper provides a holistic view of model compression, highlighting effective implementations and outlining promising future directions. To the best of our knowledge, this is the first paper that provides a focused survey of LLM compression techniques from the lens of resource-constrained environments.

The paper is organized as follows: Section II outlines the challenges that necessitate LLM compression; Section III provides an in-depth examination of compression techniques, going through the underlying principles, variants, and effective examples. Section IV discusses promising areas of research within model compression and Section V provides a conclusion.

## II. Challenges in using Large Language Models

To realize the potential of LLMs in mobile, edge, and IoT sensors, several key challenges must be overcome. Key challenges currently being addressed include:

- **Computational Cost**: LLMs require significant computational cost for inference. According to [2] the inference cost of GPT-3 2.7B model is 2 FLOPS (Floating point operations per second) per active parameter per token, bringing up the cost per token to 5.4 GigaFLOPs. This computational cost is often prohibitive to run the models directly on mobile and edge devices.

- **Memory**: LLMs, with millions or billions of parameters, often require gigabytes of GPU memory for serving. The memory requirements are a factor of the number of parameters, the number of bytes required per parameter and additional memory required (overhead factor). Even a 7B parameter model at 16-bit precision requires over 16 GB of memory [3]. This is usually well-beyond what is available in most mobile and edge devices.
- **Energy Consumption and Latency Requirements**: The high computation cost for serving LLMs also leads to significant energy usage and high-latency. Mobile devices have limited battery capacity and strict latency-requirements for applications. Thus, using LLMs on-device without any compression is impractical.
- **Deployment Complexity**: Deployment is complicated by the heterogeneity of edge and mobile hardware. While LLMs are optimized for special and general-purpose GPUs, they may need additional tuning for specific hardware targets.

In this paper, we will address ways to make LLMs and deep learning models smaller, thereby reducing their computation requirements and memory footprint, while making inference faster.

## III. Way to adapt Large Language Models for Edge Devices - Key Methods and Techniques

### A. Knowledge Distillation

Knowledge Distillation (KD) is a model compression technique where, generally, a smaller student model is trained to mimic a larger teacher model. The idea was first introduced by [12] for reducing the size of an ensemble of models, and then generalized by [11] and [38]. The key idea here is that the knowledge learned by a large complex model (teacher model) can be effectively learned by a smaller, simpler model (student model) through the process of distillation. The student model is optimized to match the teacher's outputs in addition to the ground truth labels. Therefore, the student learns to reproduce the teacher's function with high accuracy, with much fewer parameters. This enables student models to be deployed on resource constrained environments like mobile phones and edge devices.

#### 1) The Basics of Knowledge Distillation

In Knowledge Distillation, the student model is trained to learn the feature predictions from the teacher. In order to do that, a fully-trained teacher model is used to run inference on a dataset (sometimes referred to as a transfer dataset) and the predictions are called soft targets. The teacher model uses a high value of the temperature parameter T resulting in a smoother probability distribution over classes, which is more informative than binary true labels [13] or only the most-likely next token in case of LLMs. The student model predictions also use the same value of temperature during training. The cross-entropy loss (or Kullback-Leibler divergence loss) between the student model predictions and the soft targets, called the distillation loss, is minimized.

The loss function also consists of a second component which minimizes the loss between student model predictions and the ground truth labels (hard targets), called the student loss. For this component, the student model temperature T is set to 1. Hinton et al. [11] uses a weighted sum of the two losses to train the student model. The loss function becomes:

$$\text{Combined loss} = \alpha \cdot L_{CE}(y, \sigma(z_s; T=1)) + (1-\alpha)T^2 \cdot L_{CE}(\sigma(\tfrac{z_T}{T}), \sigma(\tfrac{z_s}{T})) \quad (1)$$

Where, $L_{CE}$ = Crossentropy loss; y = True labels; $z_T$ = Teacher model logits; $z_S$ = Student model logits; $\sigma$ = Softmax function; $\alpha$ = Weighting coefficient; T = Temperature.

The distillation loss is multiplied by $T^2$ to appropriately scale its gradients.

#### 2) Different Forms of Knowledge Distillation

##### a) Soft-Target Distillation (Logit-based Distillation)

Soft target, or soft label, distillation is the most popular form of KD [11] and the one described in the previous section. Soft targets are predictions from the teacher model and act as both labels to train the student model on and regularizers, as the higher temperature leads to label smoothening. Logit-based distillation is similar to soft target distribution, with the logits from the teacher and student models used directly, without applying the softmax function on them. The blue box in Fig.1 shows soft-target distillation and its two loss components.

##### b) Feature-Based Distillation

Feature-based KD uses information from the intermediate layers in the network to distill knowledge between the teacher model and the student model. This idea was introduced by [15] in the FitNets paper, where activations from specific hidden layers in the teacher model, termed as hints, were used to train corresponding layers in a thinner student model. The model used an intermediate regressor to convert from the student model hidden representation space to the teacher model hidden representation space, and used a two-stage training process. Several subsequent works have expanded upon feature-based distillation, but open problems like choosing the right layers in the teacher and student models to distill and the difference in feature representations in the models still remain. The yellow box in Fig.1 depicts where feature-based distillation takes place.

##### c) Relation-Based Distillation

Unlike the previous methods which match the outputs or activations from intermediate layers between the teacher and student models, relation-based distillation tries to preserve the relationship between data points or between intermediate representations. In the Relational Knowledge Distillation paper [16] the student model learns to align pairwise distances between data points in the latent space (called distance-wise loss) with the teacher's pairwise distances. Additionally, angles formed by triplets of points (called angle-wise loss) are also aligned. The red box in Fig 1. shows relation-based distillation based on [16].

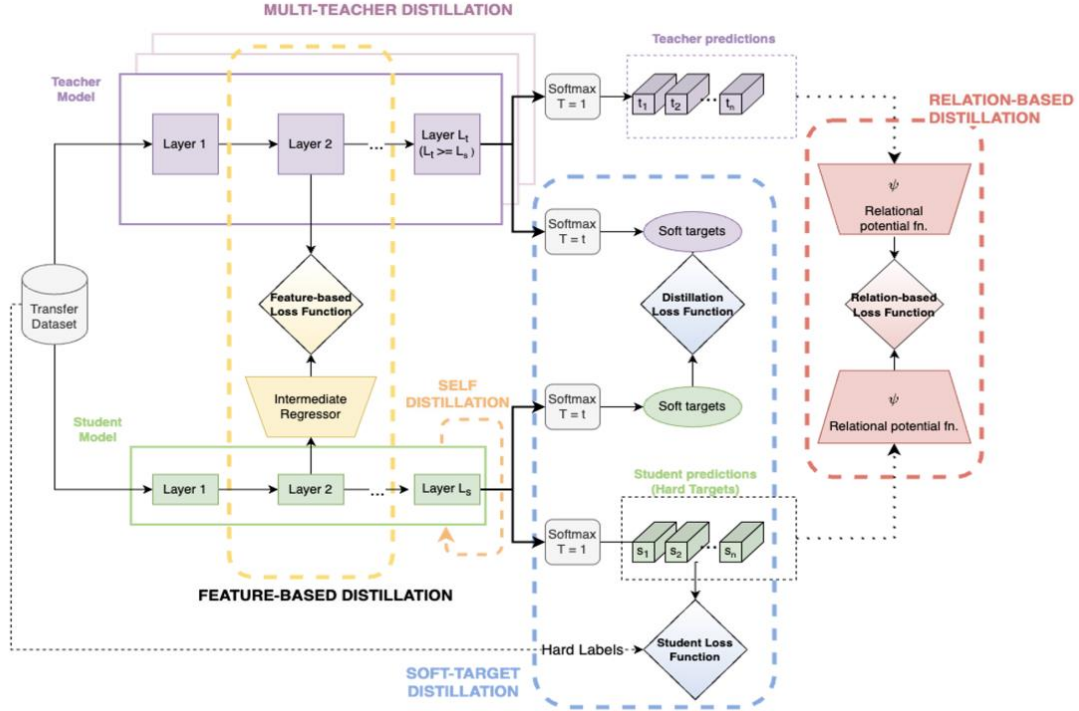

Fig 1. Different forms of Knowledge Distillation. The student model is represented by the green box and the teacher model is represented by the purple box. The outputs of the two models are used for soft-target distillation; pairs/triplets of outputs are used for relation-based distillation; the latent/feature-space representations are used for feature-based distillation; multiple teacher models are used for multi-teacher distillation; the student model itself is used for self-distillation.

*d) Self-Distillation*

In self-distillation, a model learns from itself or from models with the same architecture. This can be done in a few ways: i) the later layers of a model can be used to teach the earlier layers [17] ii) knowledge from earlier versions of the model can be used to train a later version [18]. In this method, a large teacher model is not required to train the student model. Self-distillation also effectively regularizes the model by smoothing its own predictions, which can lead to higher accuracy. This permits smaller model architectures to perform better, without the need for larger teacher models, making it a particularly useful technique for smaller models that can be used in resource constrained settings.

*e) Multi-Teacher Distillation*

In multi-teacher distillation, the student model learns from multiple teacher models simultaneously. The different teacher models may have been trained on heterogeneous datasets, and therefore may excel in different tasks. Furthermore, the distillation can happen through a combination of output (soft target) distillation and relational distillation with multiple teachers [19].

*3) Examples of Knowledge Distillation*

Notable examples of KD are summarized in TABLE I.

### B. Model Quantization

Model Quantization reduces the numerical precision, i.e., number of bits required, for the model's parameters. A trained model consists of its graph structure and parameters (weights and activations). For a non-quantized model, the parameters are usually in the form of 32-bit (FP32) or 16-bit floats (FP16, BF16). Quantization reduces the numerical precision of these weights and, often, the activations, by storing them as 8-bit integers, 4-bit integers, or even single bits [36]. The lower bit-precision leads to a reduced memory footprint for the model (by 4x, 8x or even 32x compared to the FP32 representation) and speeds up inference as integer arithmetic can be more efficient than floating-point operations. This has the added benefit of reducing the energy required to serve the models. Quantization to INT8 or lower precision also enables deployment on hardware optimized for or restricted to integer operations. The primary trade-off of quantization is a potential drop in accuracy due to the reduced precision, but various techniques can minimize this loss. In the following section we discuss different quantization strategies and a few applications of quantization.

*1) Post-Training Quantization*

Post-training Quantization (PTQ) converts a pre-trained full-precision model to a low-precision model after the training process is complete. PTQ is fast and doesn't require retraining the model. The simplest form of PTQ directly quantizes the weights after training by mapping FP32 to lower precision, but this can result in accuracy loss. More sophisticated PTQ methods have been developed that use a calibration dataset to determine the scale and zero-point for mapping floating-point values to lower precision. The uniform affine quantizer and uniform symmetric quantizer [25] are examples of calibration-based PTQ. Such techniques have been effective in reducing the precision to INT8 with minimal reductions in accuracy. However, reducing the precision even further requires more framework to decompose the quantization of weights into ternary optimization problems which can then be recombined to form INT3 values. Nagel et al [27] proposed an adaptive rounding strategy in PTQ to round weights based on their impact on the overall model accuracy and was able to quantize several CNN models to INT4 precision within minimal reduction in accuracy. This was also achieved in BrecQ [28] where a layer-wise block reconstruction strategy was used to quantize the layers of models.

TABLE I. EXAMPLES OF KD ON LLMS

| Paper | High-level Approach | Teacher Model(s) | Student Models | Notable Results |
|---|---|---|---|---|
| GKD [20] | On-policy learning: student model generates its own output sequences used for KD | T5-XL (3B params) | T5-Small (77M params), T5-Base (250M params), T5-Large (800M params) | Smallest student model surpasses few-shot performance larger PaLM (540B) for machine translation. |
| Mini LLM [21] | Reverse KLD Loss, policy gradient optimization | GPT-2 XL (1.5B); OPT (13B) [14]; LLaMA (13B) | GPT-2 (120M, 340M, 760M); OPT (1.3B, 2.7B, 6.7B); LLaMa (7B) | Outperforms standard KD using forward KLD loss across various tasks. |
| DISTI LLM [22] | Skewed KLD loss, adaptive off-policy learning | T5-XL (3B); GPT-2 1.5B; OPT-13B; LLaMA-13B. | T5 (Small, Base, Large); GPT-2 (120M, 340M, 760M); OPT (1.3B, 2.7B, 6.7B); LLaMA 7B, | Outperforms standard KD across summarization, translation, reasoning. Shows 2.5x-4.3x training speedup compared to on-policy KD. |

*2) Quantization-Aware Training*

Quantization-aware training (QAT) [29] simulates quantization during training time and allows the model to learn parameters that are more robust to the effects of quantization. This results in higher accuracy than PTQ. In QAT, the forward pass has simulated quantization operations added to the standard operations. In addition to the full-precision calculations, weights and activations are quantized and used to compute the output of each layer. Since quantization operations are non-differentiable, a straight-through estimator (STE) [31] is used to compute the gradients for the quantized parameters. The STE passes through the gradient i.e., it acts as an identity function. The full-precision gradient calculation happens normally, through the chain-rule and is used to update the weights.

QAT generally achieves much higher accuracy than PTQ, particularly for lower numerical precision levels like INT8 and INT4. It is also more stable to the errors introduced in the quantization process. But this comes at the cost of increased training time compared to PTQ, which doesn't need retraining. QAT usually requires lower learning rates and adds extra computation to the forward pass of training. QAT is also more complex to implement and has more hyperparameters to tune.

*3) A Few Special Cases of Quantization*

  *a) Mixed Precision Quantization*

Mixed precision allows for different parts of a model to be quantized with different bit-precision [32]. This provides more fine-grained control over the efficiency vs accuracy tradeoff. Different layers of a transformer model may have different sensitivities to quantization. While the final output and bottleneck layers require higher bit-precision (FLOAT-16 or higher) after quantization, other layers like layer normalization or intermediate attention output layers could be quantized to INT8 or even INT4 precision. ZeroQuant [34] provides an example of mixed precision quantization where the fully-connected modules were quantized to INT4, while the attention weights and activations were quantized to INT8 on BERT and GPT-style models, resulting in a 3x reduction in memory footprint compared to an FP16 model.

Bondarenko et al. [33] extends mixed-precision quantization further by introducing per-embedding-group quantization where bit-precision can vary based on embedding groups. Different embedding groups within a transformer may use different levels of quantization, based on their relative importance to the model. Mixed precision quantization reduces accuracy degradation significantly compared to uniform precision quantization while also making it easier to tailor the model to compute-limited surfaces that it may be served on. While identifying the ideal quantization level for each layer is an open area of research, reinforcement learning and neural architecture search-based approaches have had some success [35].

  *b) Binary and Ternary Quantization*

While typical quantization strategies reduce model parameters to FP16, INT8 or INT4, binary and ternary quantization take it even further. Binary quantization [24] uses 1-bit per parameter, so the possible values are +1 and -1. This approach was pioneered by BinaryConnect [36]. Ternary quantization uses 2-bits per parameter, but only three distinct values of +1, 0 and -1 [37]. The inclusion of the zero allows the network to completely mask out weights that are not important and can lead to better accuracy than binary quantization. These approaches can result in a 32x to 16x reduction in model size compared to a standard FP32 parameter model.

*4) Distillation and Quantization*

While KD and quantization have been discussed separately, they are often used together to achieve compounding gains when reducing model size. KD helps smaller models learn more efficiently from larger models which helps maintain accuracy of quantized models, while quantization reduces the student model's bit-precision to make the model even smaller. Polino et al. [39] had proposed quantized distillation which introduced distillation loss into the training of a reduced precision student model, which helps it learn from a larger full-precision teacher. They also introduced differentiable quantization which helps converge to the optimal location of quantization points through stochastic gradient descent. ZeroQuant [34], discussed earlier, also had a version of the model where quantization was used along with a novel version of distillation called Layer-by-layer Knowledge Distillation (LKD) to reduce the model to INT4 and INT8 mixed precision.

*5) Examples of Model Quantization*

Notable examples of Model Quantization are summarized in TABLE II.

## C. Model Pruning

Model or network pruning is a technique of removing redundant or low-importance components of a model to reduce its overall size. The components that are removed could be neurons, weights, attention heads, filters (in CNNs) or even entire layers. The objective of Model Pruning is to remove nodes and connections that contribute very little to the model's output. The pruned model can be significantly smaller than the full model, making it faster and much cheaper for inference, while using less energy. This makes the model more suitable for mobile devices and low power settings. Pruning may also result in better generalization and makes further fine-tuning faster.

The idea of Model Pruning is not new. The Optimal Brain Damage paper [44] in 1989 used second derivative information to decide which weights in a model can be pruned. The Optimal Brain Surgeon [45] improved upon this idea in 1993 with a more accurate estimate of the effect of removing weights. More recently, Han et al. [46] introduced a three-step process: i) training networks to learn which connections are important ii) pruning the unimportant connections iii) retrain network to fine tune the weights of the remaining connections. This work focuses on learning which connections are important in a network and removing the low-weight connections, followed by pruning neurons that have zero input or output connections. This led to much sparser models, with a 9x and 13x reduction in model parameters for the AlexNet and VGG-16 models, respectively, without any loss in accuracy.

Pruning techniques can be divided into two broad categories based on what gets pruned:

- **Unstructured**: The general case of non-important weights and neurons being removed is a form of unstructured pruning [44], [45], [46]. This yields sparse weight matrices and can usually be done at a more granular level. The disadvantage of unstructured pruning is that the hardware may not be able to take full advantage of the sparsity, as GPUs are optimized for dense matrix operations.
- **Structured**: In structured pruning, whole components like layers, filters or channels are removed [54]; or constraints like N:M sparsity [60], block sparsity or vector sparsity are enforced on the weights. This approach is more hardware friendly, but the degree of pruning may be lesser if accuracy needs to be preserved.

The rationale behind pruning was strengthened by the Lottery Ticket Hypothesis [43] paper where the authors were performance to the original networks through iterative pruning. In this work, the pruning led to "winning tickets" which are smaller networks that learn faster than the original network while reaching equal or higher test accuracy and generalizing better. In their experiments, the authors were able to reduce model parameters up to even 90% with very little effect on accuracy. Other interesting advances in pruning come from the use of reinforcement learning for pruning and movement-based pruning, geared towards transfer learning and fine-tuning. He et al. proposed AutoML [48] for Model Compression (AMC) which uses reinforcement learning to automatically determine the sparsity for each layer and then perform pruning based on the sparsity. The paper also provides the option of using different reward schemes which can determine whether the compression is optimizing for resource-constrained environments or accuracy-guarantees. In Movement Pruning [47], the change of weights during fine-tuning, i.e. first-order information, is taken into consideration to determine the weights to prune. In their experiments, movement-based pruning outperforms more common approaches like magnitude-based and regularization-based pruning.

TABLE II. EXAMPLES OF MODEL QUANTIZATION

| Paper | Quantization Approach | Target Model | Bit-width Post-Quantization | Notable Results |
|---|---|---|---|---|
| **Zero Quant [34]** | PTQ; Mixed Precision | BERT, GPT-J (6B), GPT-NeoX (20B) | All INT8 (W8A8); Mixed W4/W8 A8 | 2x to 3x memory reduction; Minimal drop in accuracy; ~5.2x faster inference |
| **GPTQ [30]** | PTQ | OPT-175B [14], BLOOM-176B | Weights INT3-4 (W3-4), Activation FP16 (A16) | 4x to 5x memory reduction; 3.25x to 4.5x faster inference; negligible accuracy drop |
| **Smooth Quant [40]** | PTQ + Activation smoothing | OPT-175B, BLOOM-176B, GLM-130B, MT-NLG 530B, Llama-2 (70B), Falcon, Mistral, Mixtral | All INT8 (W8A8) | 2x memory reduction compared; 1.51x to 1.56x faster inference; Works on 530B param model; Accuracy close to FP16 model. |
| **Zero Quant V2 [43]** | PTQ | OPT Family (125M to 175B) [14]; BLOOM Family (560M to 176B) [60] | W8A16; W4A16; W4A8 | ~2x to ~4x memory reduction; Low-Rank Compression for accuracy recovery |

While pruning is an effective technique to reduce model size, it has a few limitations when it comes to real-world deployment. Weight pruning produces sparse weight matrices, but most GPUs are optimized for dense matrix operations, so the increase in sparsity may not translate to improvements in inference speed if the hardware cannot take advantage of it. Specialized libraries are available to take advantage of this sparsity like Nvidia's cuSparse, TensorRT, but they are not optimized for or widely used for deep learning inference. Pruning, especially iterative pruning, also adds to the compute required post-training. Determining what to prune is non-trivial, and while there are promising areas of research looking into this, it remains an open problem.

TABLE III. COMPARISON OF COMPRESSION TECHNIQUES.

| Technique | Primary Goal | Advantages | Disadvantages | Memory Reduction | Computation Cost Reduction | Accuracy Degradation | Hardware Adaptability (of the final model) | Edge Suitability |
|---|---|---|---|---|---|---|---|---|
| Knowledge Distillation | Transfer knowledge from larger, complex teacher model to a smaller, more efficient student model | Improved accuracy, faster inference, better generalization for smaller student model | Requires larger teacher model, needs to be trained and higher training cost | Moderate to High | Moderate to High | Often improves accuracy | High | High |
| Quantization | Reduce numerical precision by mapping FP32 parameters to lower bit-widths (e.g.: FP16, INT8, INT4) | Reduced model size, faster inference, lower power, hardware compatibility | Potential accuracy loss, calibration for PTQ, additional training complexity for QAT | High to Very High | Moderate to Very High | Very Low (QAT) to High (PTQ) | Very high | Very High |
| Pruning | Reduce parameter count by removing weights, connections, neurons, filters, etc. | Reduced model size, potentially faster inference, improved generalization | Unstructured pruning is hard to accelerate; complexity in choosing what to prune | Moderate (structured) to Very High (unstructured) | Low to Very High | Very Low to Medium | Low (unstructured) to High (structured) | Moderate (unstructured) to High (structured) |

*D. Other Strategies to Improve Computational Efficiency*

While not considered as model compression techniques, there are a few techniques which help improve computational efficiency. We briefly discuss some of these techniques here:

*1) Mixture of Experts*

Mixture of Experts (MoEs) divides the model into multiple sub-models called experts and uses a gating mechanism (router) to route each request to only a small number of experts [49]. MoEs replace the dense feed-forward networks with multiple parallel expert networks, only a few of which are activated at a time during training and inference, which brings down the computation cost. The gating mechanism, called the router, is also a network that learns which experts to route a request to, based on the input features. Rather than allow for smaller models, MoEs permit models to be larger for the same amount of compute per request so its primary goal differs from traditional compression techniques focused solely on parameter reduction. MoEs have been used to scale up model capacity by 1000x without an increase in computational cost in [50]. In GLaM [51], sparsely activated MoE architecture was used to scale LLMs to 1.2 trillion parameters or 7x the number of parameters as GPT-3, with 51.4% of the computational cost and only 35.4% of the energy cost.

*2) Early-Exit Strategies*

A big contributor to the computational cost of transformer-based and deep learning models are the number of layers in the forward pass. Models progressively learn better feature representations with more layers, with earlier layers learning simpler feature representations and later layers learning more complex features. However, the inputs to the model vary in terms of difficulty to predict, and some inputs can be predicted accurately with the features learnt in the earlier layers of the model. Early-exit strategies take advantage of this fact by providing a way to make predictions without going through all the layers in the model. These strategies augment models with side branch classifiers/outputs at earlier layers in the network, which can be used selectively. When outputs are made from these early-exit branches, a large part of the computation can be skipped, making the inference faster and more efficient. This idea was introduced in the BranchyNet paper [52] where early exit branches were added to popular architectures in the literature (LeNet, AlexNet, ResNet) to exit early when an entropy threshold was met. This resulted in a 2x-6x speed up on CPU and GPU.

IV. FUTURE DIRECTIONS AND AREAS OF RESEARCH IN MODEL COMPRESSION

The field of model compression is still evolving with more sophisticated techniques and hybrid approaches being developed to get the best out of LLMs. Some of the advancements are:

*A. Newer forms of Knowledge Distillation*

New forms of KD have emerged like i) Fine-tune-CoT (Chain of thought) [56] to transfer reasoning capabilities from >100B parameter LLMs to much smaller student models ii) Distilling Step-by-Step [57] which extracts LLM rationales for additional supervision during fine-tuning or distillation. In these methods, the student model is trained not just on the final outputs, but also on the intermediate steps. There is also ongoing research to reduce the distribution skew between the training and inference distributions by using techniques like on-policy distillation [20] where the student model is trained using its own generated outputs.

## B. Smaller Float Representations

Newer forms of floats like FP8 [61] and NF4 (NormalFloat 4-bit) [58] have been introduced in the last few years which enable models to use lower bit-widths while maintaining acceptable accuracy and accelerating both training and inference. Reference [61] introduced FP8 with two formats (E4M3 and E5M2) and discussed the tradeoff between the two formats. The paper demonstrated experiments where the weights, activations and gradients of FP16 models were clipped to FP8 and the model accuracy remained comparable to the FP16 baselines. NF4 was designed to be optimal for quantizing data that follows a normal distribution. There is also growing hardware support for FP8 with Nvidia's Hopper Architecture (H100/H200) and AMD Instinct GPUs supporting it natively.

## C. Neural Architecture Search (NAS)

Neural architecture search automates the process of designing the neural architecture, with a customizable search objective that is capable of incorporating efficiency metrics. This is being extended to LLMs in works such as [59] where NAS is used to find less computationally complex architectures on LLMs with over a Billion parameters. NAS, particularly hardware-aware NAS, can lead to models that are smaller, faster and more efficient while achieving compression by design.

## D. Outlier-based Quantization and Low Bit-Precision Matrix Multiplications

Quantization and matrix multiplication schemes that work on INT8 and INT4 precision for most of the calculations, but treat only the outliers or salient weight with higher precision have been introduced in [23] and [42]. These schemes have no performance degradation compared to the full-precision equivalents. LLM.int8() [23] enabled OPT-175B/BLOOM to be used on a single GPU server, while AWQ [42] enabled the deployment of the 70B Llama-2 model on mobile GPUs.

## E. Quantized Finetuning

Quantization can also be applied to LLM finetuning, in addition to training. This was demonstrated by QLoRA [58], where LLMs were finetuned with significantly lower memory requirements by using the NF4 float representation, permitting the finetuning of a 65 billion parameter model on a single 48GB GPU, while maintaining full 16-bit finetuning performance.

## V. CONCLUSION

In this paper we have discussed three important techniques for model compression: KD, Quantization and Pruning, and its different forms in detail. These techniques were compared based on their memory and computational cost reduction, accuracy degradation, hardware and edge suitability. We have listed out examples from literature where these techniques were effectively applied on LLMs to reduce model size and computational cost, accelerate inference and improve energy efficiency. We also discussed a few additional techniques which can improve computational efficiency and the promising areas of ongoing research and future directions in the field. We hope this work proves to be a useful reference for researchers and practitioners in the field to apply model compression to their own work, and further the research in this exciting domain.

## REFERENCES

[1] G. Team et al., "Gemini: A Family of Highly Capable Multimodal Models," Jun. 17, 2024, arXiv: arXiv:2312.11805. doi: 10.48550/arXiv.2312.11805.

[2] T. Brown et al. "Language models are few-shot learners." Advances in neural information processing systems 33 (2020): 1877-1901.

[3] H. Touvron et al., "LLaMA: Open and Efficient Foundation Language Models," Feb. 27, 2023, arXiv: arXiv:2302.13971. doi: 10.48550/arXiv.2302.13971.

[4] A. Grattafiori et al., "The Llama 3 Herd of Models," Nov. 23, 2024, arXiv: arXiv:2407.21783. doi: 10.48550/arXiv.2407.21783.

[5] A. Fan, M. Lewis, and Y. Dauphin, "Hierarchical Neural Story Generation," May 13, 2018, arXiv: arXiv:1805.04833. doi: 10.48550/arXiv.1805.04833

[6] R. Thoppilan et al., "LaMDA: Language Models for Dialog Applications," Feb. 10, 2022, arXiv: arXiv:2201.08239. doi: 10.48550/arXiv.2201.08239.

[7] M. U. Hadi et al., "A Survey on Large Language Models: Applications, Challenges, Limitations, and Practical Usage," Jul. 10, 2023. doi: 10.36227/techrxiv.23589741.v1.

[8] J. Jumper et al., "Highly accurate protein structure prediction with AlphaFold," Nature, vol. 596, no. 7873, pp. 583–589, Aug. 2021, doi: 10.1038/s41586-021-03819-2.

[9] X. Zhu, J. Li, Y. Liu, C. Ma, and W. Wang, "A Survey on Model Compression for Large Language Models," Transactions of the Association for Computational Linguistics, vol. 12, pp. 1556–1577, 2024, doi: 10.1162/tacl_a_00704.

[10] Y. Tang et al., "A Survey on Transformer Compression," Apr. 07, 2024, arXiv: arXiv:2402.05964. doi: 10.48550/arXiv.2402.05964.

[11] G. Hinton, O. Vinyals, and J. Dean, "Distilling the Knowledge in a Neural Network," Mar. 09, 2015, arXiv: arXiv:1503.02531. doi: 10.48550/arXiv.1503.02531.

[12] C. Buciluă, R. Caruana, and A. Niculescu-Mizil, "Model compression," in Proceedings of the 12th ACM SIGKDD international conference on Knowledge discovery and data mining, Philadelphia PA USA: ACM, Aug. 2006, pp. 535–541. doi: 10.1145/1150402.1150464.

[13] J. Gou, B. Yu, S. J. Maybank, and D. Tao, "Knowledge Distillation: A Survey," Int J Comput Vis, vol. 129, no. 6, pp. 1789–1819, Jun. 2021, doi: 10.1007/s11263-021-01453-z.

[14] S. Zhang et al., "OPT: Open Pre-trained Transformer Language Models," Jun. 21, 2022, arXiv: arXiv:2205.01068. doi: 10.48550/arXiv.2205.01068.

[15] A. Romero, N. Ballas, S. E. Kahou, A. Chassang, C. Gatta, and Y. Bengio, "FitNets: Hints for Thin Deep Nets," Mar. 27, 2015, arXiv: arXiv:1412.6550. doi: 10.48550/arXiv.1412.6550.

[16] W. Park, D. Kim, Y. Lu and M. Cho, "Relational knowledge distillation." Proceedings of the IEEE/CVF conference on computer vision and pattern recognition. 2019.

[17] L. Zhang, J. Song, A. Gao, J. Chen, C. Bao and K. Ma, "Be Your Own Teacher: Improve the Performance of Convolutional Neural Networks via Self Distillation," 2019 IEEE/CVF International Conference on Computer Vision (ICCV), 2019, pp. 3712-3721, doi: 10.1109/ICCV.2019.00381.

[18] T. Furlanello, Z. Lipton, M. Tschannen, L. Itti, A. Anandkumar, "Born again neural networks", InInternational conference on machine learning 2018 Jul 3 (pp. 1607-1616). PMLR.

[19] S. You, C. Xu, C. Xu, and D. Tao, "Learning from Multiple Teacher Networks," in Proceedings of the 23rd ACM SIGKDD International Conference on Knowledge Discovery and Data Mining, Aug. 2017, pp. 1285–1294. doi: 10.1145/3097983.3098135.

[20] R. Agarwal et al., "On-policy distillation of language models: Learning from self-generated mistakes," in the Twelfth International Conference on Learning Representations 2024 May.

[21] Y. Gu, L. Dong, F. Wei, and M. Huang, "MiniLLM: Knowledge Distillation of Large Language Models," Apr. 10, 2024, arXiv: arXiv:2306.08543. doi: 10.48550/arXiv.2306.08543.

[22] J. Ko, S. Kim, T. Chen, and S.-Y. Yun, "DistiLLM: Towards Streamlined Distillation for Large Language Models," Jul. 03, 2024, arXiv: arXiv:2402.03898. doi: 10.48550/arXiv.2402.03898.

[23] T. Dettmers, M. Lewis, Y. Belkada, and L. Zettlemoyer, "GPT3.int8(): 8-bit Matrix Multiplication for Transformers at Scale," Advances in Neural Information Processing Systems, vol. 35, pp. 30318–30332, Dec. 2022.

[24] H. Wang, et al. "Bitnet: Scaling 1-bit transformers for large language models." arXiv preprint arXiv:2310.11453 (2023).

[25] R. Krishnamoorthi, "Quantizing deep convolutional networks for efficient inference: A whitepaper," Jun. 21, 2018, arXiv: arXiv:1806.08342. doi: 10.48550/arXiv.1806.08342.

[26] P. Wang, Q. Chen, X. He, and J. Cheng, "Towards Accurate Post-training Network Quantization via Bit-Split and Stitching," in Proceedings of the 37th International Conference on Machine Learning, PMLR, Nov. 2020, pp. 9847–9856

[27] M. Nagel, R. A. Amjad, M. Van Baalen, C. Louizos and T. Blankevoort, "Up or down? Adaptive rounding for post-training quantization", Proceedings of the 37th Int. Conference on Machine Learning, pp. 7197-7206, 2020.

[28] Y. Li et al., "BRECQ: Pushing the Limit of Post-Training Quantization by Block Reconstruction," Jul. 25, 2021, arXiv: arXiv:2102.05426. doi: 10.48550/arXiv.2102.05426.

[29] B. Jacob et al., "Quantization and Training of Neural Networks for Efficient Integer-Arithmetic-Only Inference," in 2018 IEEE/CVF Conference on Computer Vision and Pattern Recognition, Jun. 2018, pp. 2704–2713. doi: 10.1109/CVPR.2018.00286.

[30] E. Frantar, S. Ashkboos, T. Hoefler, and D. Alistarh, "GPTQ: Accurate Post-Training Quantization for Generative Pre-trained Transformers," Mar. 22, 2023, arXiv: arXiv:2210.17323. doi: 10.48550/arXiv.2210.17323.

[31] Y. Bengio, N. Léonard, and A. Courville, "Estimating or Propagating Gradients Through Stochastic Neurons for Conditional Computation," Aug. 15, 2013, arXiv: arXiv:1308.3432. doi: 10.48550/arXiv.1308.3432.

[32] A. Gholami, S. Kim, Z. Dong, Z. Yao, M. W. Mahoney, and K. Keutzer, "A Survey of Quantization Methods for Efficient Neural Network Inference," In Low-power computer vision (pp. 291-326).

[33] Y. Bondarenko, M. Nagel, and T. Blankevoort, "Understanding and Overcoming the Challenges of Efficient Transformer Quantization," Sep. 27, 2021, arXiv: arXiv:2109.12948. doi: 10.48550/arXiv.2109.12948.

[34] Z. Yao, R. Yazdani Aminabadi, M. Zhang, X. Wu, C. Li, and Y. He, "ZeroQuant: Efficient and Affordable Post-Training Quantization for Large-Scale Transformers," Advances in Neural Information Processing Systems, vol. 35, pp. 27168–27183, Dec. 2022.

[35] K. Wang, Z. Liu, Y. Lin, J. Lin, and S. Han, "HAQ: Hardware-Aware Automated Quantization With Mixed Precision," in 2019 IEEE/CVF Conference on Computer Vision and Pattern Recognition (CVPR), Long Beach, CA, USA: IEEE, Jun. 2019, pp. 8604–8612. doi: 10.1109/CVPR.2019.00881.

[36] M. Courbariaux, Y. Bengio, and J. P. David, "Binaryconnect: Training deep neural networks with binary weights during propagations," in J. P. in Advances in Neural Information Processing Systems 2015, 28.

[37] B. Liu, F. Li, X. Wang, B. Zhang, and J. Yan, "Ternary Weight Networks," in ICASSP 2023 - 2023 IEEE International Conference on Acoustics, Speech and Signal Processing (ICASSP), Jun. 2023, pp. 1–5. doi: 10.1109/ICASSP49357.2023.10094626.

[38] L. J. Ba and R. Caruana, "Do Deep Nets Really Need to be Deep?," in Advances in Neural Information Processing Systems 2014, 27.

[39] A. Polino, R. Pascanu, and D. Alistarh, "Model compression via distillation and quantization," Feb. 15, 2018, arXiv: arXiv:1802.05668. doi: 10.48550/arXiv.1802.05668.

[40] G. Xizao, J. Lin, M. Seznec, H. Wu, J. Demouth, and S. Han, "SmoothQuant: Accurate and Efficient Post-Training Quantization for Large Language Models," In International Conference on Machine Learning (pp. 38087-38099). PMLR.

[41] Z. Yao, X. Wu, C. Li, S. Youn, and Y. He, "ZeroQuant-V2: Exploring Post-training Quantization in LLMs from Comprehensive Study to Low Rank Compensation," May 26, 2023, arXiv: arXiv:2303.08302. doi: 10.48550/arXiv.2303.08302.

[42] J. Lin, J. Tang, H. Tang, S. Yang, G. Xiao, and S. Han, "AWQ: Activation-aware Weight Quantization for On-Device LLM Compression and Acceleration," GetMobile: Mobile Comp. and Comm., vol. 28, no. 4, pp. 12–17, Jan. 2025, doi: 10.1145/3714983.3714987.

[43] J. Frankle and M. Carbin, "The Lottery Ticket Hypothesis: Finding Sparse, Trainable Neural Networks," Mar. 04, 2019, arXiv: arXiv:1803.03635. doi: 10.48550/arXiv.1803.03635.

[44] Y. LeCun, J. Denker, and S. Solla, "Optimal brain damage. Advances" In Neural Information Processing Systems, 1989.

[45] B. Hassibi, D. G. Stork, and G. J. Wolff, "Optimal Brain Surgeon and general network pruning," in IEEE International Conference on Neural Networks, San Francisco, CA, USA: IEEE, 1993, pp. 293–299. doi: 10.1109/ICNN.1993.298572.

[46] S. Han, J. Pool, J. Tran, and W. Dally, "Learning both weights and connections for efficient neural network" In the Advances In Neural Information Processing Systems, 2015.

[47] V. Sanh, T. Wolf, and A. Rush, "Movement Pruning: Adaptive Sparsity by Fine-Tuning," in Advances in Neural Information Processing Systems, Curran Associates, Inc., 2020, pp. 20378–20389.

[48] Y. He, J. Lin, Z. Liu, H. Wang, L.-J. Li, and S. Han, "AMC: AutoML for Model Compression and Acceleration on Mobile Devices," in Computer Vision – ECCV 2018, vol. 11211

[49] R. A. Jacobs, M. I. Jordan, S. J. Nowlan, and G. E. Hinton, "Adaptive Mixtures of Local Experts," Neural Computation, vol. 3, no. 1, pp. 79–87, Feb. 1991, doi: 10.1162/neco.1991.3.1.79

[50] N. Shazeer et al., "Outrageously Large Neural Networks: The Sparsely-Gated Mixture-of-Experts Layer," Jan. 23, 2017, arXiv: arXiv:1701.06538. doi: 10.48550/arXiv.1701.06538.

[51] N. Du et al. "Glam: Efficient scaling of language models with mixture-of-experts." In International conference on machine learning. PMLR, 2022.

[52] S. Teerapittayanon, B. McDanel, and H. T. Kung, "BranchyNet: Fast inference via early exiting from deep neural networks," in 2016 23rd International Conference on Pattern Recognition (ICPR), Cancun: IEEE, Dec. 2016, pp. 2464–2469. doi: 10.1109/ICPR.2016.7900006.

[53] P. Lewis et al., "Retrieval-augmented generation for knowledge-intensive nlp tasks," In Advances in neural information processing systems 33 (2020): 9459-9474.

[54] S. Anwar, K. Hwang, and W. Sung, "Structured Pruning of Deep Convolutional Neural Networks," J. Emerg. Technol. Comput. Syst., vol. 13, no. 3, pp. 1–18, Jul. 2017, doi: 10.1145/3005348.

[55] T. Liang, J. Glossner, L. Wang, S. Shi, and X. Zhang, "Pruning and quantization for deep neural network acceleration: A survey," Neurocomputing, vol. 461, pp. 370–403, Oct. 2021, doi: 10.1016/j.neucom.2021.07.045.

[56] N. Ho, L. Schmid, and S.-Y. Yun, "Large Language Models Are Reasoning Teachers," Jun. 13, 2023, arXiv: arXiv:2212.10071. doi: 10.48550/arXiv.2212.10071.

[57] C.-Y. Hsieh et al., "Distilling Step-by-Step! Outperforming Larger Language Models with Less Training Data and Smaller Model Sizes," Jul. 05, 2023, arXiv: arXiv:2305.02301. doi: 10.48550/arXiv.2305.02301.

[58] T. Dettmers, A. Pagnoni, A. Holtzman, and L. Zettlemoyer, "Qlora: Efficient finetuning of quantized llms," In Advances in neural information processing systems 36, 2023, 10088-10115.

[59] A. Sarah, S. N. Sridhar, M. Szankin, and S. Sundaresan, "LLaMA-NAS: Efficient Neural Architecture Search for Large Language Models," May 28, 2024, arXiv: arXiv:2405.18377. doi: 10.48550/arXiv.2405.18377.

[60] Y. Zhang et al., "Learning best combination for efficient n: M sparsity." In Advances in Neural Information Processing Systems 35 (2022)

[61] P. Micikevicius et al., "FP8 Formats for Deep Learning," Sep. 29, 2022, arXiv: arXiv:2209.05433. doi: 10.48550/arXiv.2209.05433.

[62] "unsloth/DeepSeek-V3-0324-GGUF · Hugging Face." Accessed: Apr. 19, 2025. [Online]. Available: https://huggingface.co/unsloth/DeepSeek-V3-0324-GG